\documentclass{article}
\usepackage{ijcai22}

\usepackage{times}

\usepackage{soul}
\usepackage{url}
\usepackage[hidelinks]{hyperref}
\usepackage[utf8]{inputenc}
\usepackage[small]{caption}
\usepackage{graphicx}
\usepackage{amsmath, amsthm, amsfonts, amssymb, mathrsfs, multirow, verbatim} 
\usepackage{algorithm, algorithmic}
\usepackage{booktabs, tabularx} 
\usepackage{setspace}
\usepackage{afterpage}

\title{The Bid Picture: Auction-Inspired Multi-player Generative Adversarial Networks Training}

\author{
Joo Yong Shim$^1$\footnote{Contact Author}\and
Jean Seong Bjorn Choe$^1$\And
Jong-Kook Kim$^1$\\
\affiliations
$^1$Korea University, Seoul, 02841, Korea\\
\emails
shimjoo@korea.ac.kr,\ garangg@korea.ac.kr,\ jongkook@korea.ac.kr
}

\begin{document}

\maketitle

\begin{abstract}
This article proposes auction-inspired multi-player generative adversarial networks training, which mitigates the mode collapse problem of GANs. Mode collapse occurs when an over-fitted generator generates a limited range of samples, often concentrating on a small subset of the data distribution. Despite the restricted diversity of generated samples, the discriminator can still be deceived into distinguishing these samples as real samples from the actual distribution. In the absence of external standards, a model cannot recognize its failure during the training phase. We extend the two-player game of generative adversarial networks to the multi-player game. During the training, the values of each model are determined by the bids submitted by other players in an auction-like process. 
\end{abstract}

\section{Introduction}
The explosive popularity of generative tasks and applications has led to remarkable progress in generative model research \cite{DALLE,diffwave,GALIP,alqahtani2021applications}. However, there still exists a problem called the generative learning trilemma \cite{trilemma}. The generative learning trilemma is an issue in which the generative learning framework cannot meet the three essential requirements of a successful generative model at once: (i) High-Quality Samples, (ii) Fast Sampling, (iii) Mode Coverage / Diversity. Generative Adversarial Networks (GANs) \cite{goodfellow} has mode collapse problem, producing only a small subset of data or the same samples repeatedly \cite{salimans2016improved,arjovsky2017towards}. This research addresses the generative learning trilemma by focusing on the problem of GANs, specifically the mode coverage issue. Specifically, it extends the two-player min-max game of GANs to the multi-player games of multiple GANs and introduces an auction-inspired auxiliary training algorithm. Note that a preliminary version of the proposed method in this paper was published in our previous paper ~\cite{ASK2023}.

GANs are usually defined as a two-player min-max game of two neural networks: a discriminator and a generator \cite{goodfellow}. These two networks are trained in an adversarial way, where the objective function is defined as:
\begin{equation}
    \begin{split}
        \min_{G}\max_{D}V(D,G) & = \mathbb{E}_{x \sim p_{data}}\left[\log D\left(x\right)\right] \\ 
        & + \mathbb{E}_{z \sim p_{z}} \left[\log (1-D\left(G(z\right))\right]
    \end{split}
\label{eq:gan_obj}
\end{equation}

, where $x \sim p_{data}$ is an image sample from the real data distribution, and $z \sim p_z$ is a prior input noise, which is generated randomly. Here, the generator $G$ tries to minimize the objective function~\eqref {eq:gan_obj}, generating realistic samples which are indistinguishable from the discriminator $D$. Whereas the discriminator $D$ tries to maximize the objective function, identifying the fake samples generated by the generator $G$ from the real data. In the optimal scenario, it reaches the equilibrium point at the global optimal state, where $p_g = p_{data}$. In this state, the generator $G$ generates perfect samples and the discriminator $D$ is unable to distinguish the real and generated distribution, i.e. $D(x) = \frac{1}{2}$.
    
The mode collapse problem usually occurs when the generator $G$ captures only a few modes of data distribution that can easily deceive the discriminator $D$. In this state, the generator $G $ captures only a portion of the actual data distribution and produces a deceptive image, which the discriminator $D$ cannot distinguish from the real data, resulting in the illusion that the optimal situation has been reached. In this state, neither the generator $G$ nor the discriminator $D$ can be improved.

The idea starts from the question: Is it possible to detect mode collapse during training, particularly whether the generator is misleading the discriminator by generating only a limited set of modes from the dataset? In the classical training process of GANs, it is difficult for a single model to realize mode collapse during the training phase. A single discriminator is the only way in which the samples of the generator can be evaluated. Likewise, the fake samples that the discriminator can source are solely those produced by its associated generator. This prompts the following question: Wouldn't it be helpful to use appropriate external references to provide hints about the performance of the generator and the discriminator? How about extending the two-player game to the multi-player game and giving them some relative values to each other? 

After all these questions, this work introduces the auction-inspired multi-player generative adversarial networks training process. This method simultaneously trains multiple GANs by using an algorithm that includes an auction-like evaluation process and auxiliary training update. In the traditional GANs training process, the discriminator tries to approximate the distance between the real distribution of the data and the sample distribution to evaluate the generator's performance. The usual distance metric is Jensen-Shannon divergence, which is minimized when the real samples and the fake samples are completely indistinguishable. However, this distance cannot be applied when evaluating generators from the other GANs. Thus, a new score metric is defined to evaluate external GANs. The concept of an auction is adopted in this scoring process, to give a proper valuation of multiple GANs since the auctions are usually used when there is uncertainty between sellers and buyers in the valuation of items\cite{auction}. The best GANs are then selected in each step using this new score metric. An additional training process is followed, where each discriminator tries to fit its output values to the best discriminator's output. It is assumed that the measurement of the best discriminator is more accurate than the others and it will give the proper reference value during the training phase, guiding them to be on the right track. 

The main contributions of the proposed method can be summarized as follows: 
\begin{itemize}
    \item This work provides a novel idea for training GANs to solve the mode collapse problem by extending a two-player game to a multi-player game. Multiple GANs are trained effectively by adopting an auction-like evaluation approach. 
    	
    \item The additional auxiliary training process is proposed to properly guide the training process of multiple GANs and prevent them from falling into failure modes. 
    
    \item The impacts of training multi-GANs by using the proposed algorithm are examined qualitatively and quantitatively.
\end{itemize}

\section{Related Works}
There have been a lot of efforts to solve the mode collapse problem of GANs. Some of the works tried to mitigate the mode collapse in GANs by improving the learning process of GANs \cite{unrolled,wgan,wgan_gp,lsgan} or by enforcing GANs to cover diverse modes \cite{infogan,modegan}. Unrolled GANs\cite{unrolled} proposed an enhanced training process for GANs to reduce mode collapse. It examines how the discriminator will be affected in the next $k$ steps to the current update of the generator and lessens the chance of the generator overfitting to a specific discriminator, which as a result, allows generators to cover diverse modes of data distribution. WGAN \cite{wgan} and LSGAN \cite{lsgan} are other notable models which reduced the mode collapse problem. They introduced different divergence metrics to stabilize the training of GANs and generate better samples. InfoGAN \cite{infogan} tried to allay the diversity issue by enforcing GANs to cover diverse modes. Unlike the classical GANs model which uses single input latent variable z, it adds latent code $c$ to the input as $(z,c)$. Then, it extends information theoretic regulation, which tries to increase the Mutual Information (MI) between the code $c$ added in the latent space and the generated sample. This helps to learn disentangled presentations, preventing mode collapse. ModeGAN \cite{modegan}, another trial to enforce diverse modes of GANs, trains generators in collaboration using encoders. It supports sample diversity by using the fact that the fake sample generated by the generator is likely to belong to the same mode when the real sample passes via the encoder-decoder. This work belongs to both approaches, in which it tries to improve the learning process of GANs by introducing an additional auxiliary training process and enforcing diverse modes by referencing multiple generators and discriminators.

This work is not the first attempt to extend the two-player game to the multi-player game \cite{mohebbi2023games}. There were previous researches that used multiple generators or discriminators \cite{ghosh2018multi,hoang2018mgan,nguyen2017dual,mclgan,albuquerque19a}. MAD-GAN \cite{ghosh2018multi} uses multiple generators that capture different modes and the discriminator is trained to identify whether the samples are real or fake and find which generator has generated the fake sample. D2GAN \cite{nguyen2017dual} extended to a three-player game of two discriminators and one generator. D2GAN uses one more discriminator that is trained by using reverse KL distance, learning to judge fake data as real. It allows parameters to adjust the quality and diversity of the generated images by reducing the KL distance and reverse KL distance by minimizing the loss function. MCL-GAN \cite{mclgan} also uses multiple discriminators where each sample is assigned to best-suited discriminators. This makes each discriminator to be the expert model for assigned samples and helps to mitigate the mode collapse problem. This work uses multiple GANs to improve mode coverage and diversity performance, but it is different in that the proposed algorithm can be easily applied without any special structural changes, and multiple models evaluate values of each other.

\begin{algorithm*}[t]
\caption{Overall Algorithm}
\label{alg:overall_algorithm}
\begin{algorithmic}[1] 
\REQUIRE Pairs of $N$ GANs $\mathcal{G}=\{(G_1,D_1),\dots,(G_N,D_N)\}$, a dataset $\mathcal{D}$ and a hyperparameter $\lambda$.
\FOR{number of training iteration epochs}
    \FORALL{$G, D \in \mathcal{G}$}
    \STATE \textbf individual training
    \ENDFOR
    \STATE  \textbf{do} Auction and \textbf{select} the $BestGAN (G^*, D^*)$
    \FORALL{$G, D \in \mathcal{G}$}
        \FOR{Minibatch of data $x$ from $\mathcal{D}$}
            \STATE Draw a batch of samples $x'$ from $G$.
            \STATE Calculate the discriminator loss $L_D$ from $x$ and $x'$.
            \STATE Calculate the auxiliary loss $L_\text{aux}$ from $L_D$, $x$, $x'$ and $D^*$.
            \STATE Update $D$ to minimize $\mathcal{L}_D=L_D+\lambda L_{D^*}$.
            \STATE Update $G$ using $D$.
        \ENDFOR
    \ENDFOR
\ENDFOR
\end{algorithmic}
\end{algorithm*}

\section{Methods}

A set of $N$ pairs of GANs consists of a generator (G) and a discriminator (D), $\{(G_1, D_1), (G_2, D_2), \dots, (G_N, D_N)\}$ is considered. Each pair of GANs is trained to maximize performance by an auction-based two-stage update algorithm. The overall algorithm is described in Algorithm \ref{alg:overall_algorithm}.

The first-step update is conducted independently for each model, a process termed ``individual training". This update process is applied across all pairs of GANs in the same way as the standard GANs training.  For instance, when using vanilla GANs, each GANs is updated based on the objective function Eq.~\eqref{eq:gan_obj}. Note that various types of GANs can be applied by simply employing the training method and objective function for selected GANs models or algorithms.

The second-step update is ``auxiliary training", which uses the external value acquired from the best GANs model. The best GANs pair is selected using the auction-inspired valuation process. It is expected the performance of the GANs can be calibrated during auxiliary training, by calculating the relative loss that refers to the outputs of the best GANs.

\subsection{Auction-Inspired Valuation Process}
In the auction, all generators become auctioneers and present a set of generated images as a lot for the auction, and all discriminators become participants and submit bids for images. 

\begin{figure}[h]
    \centering
    \includegraphics[width=0.99\linewidth]{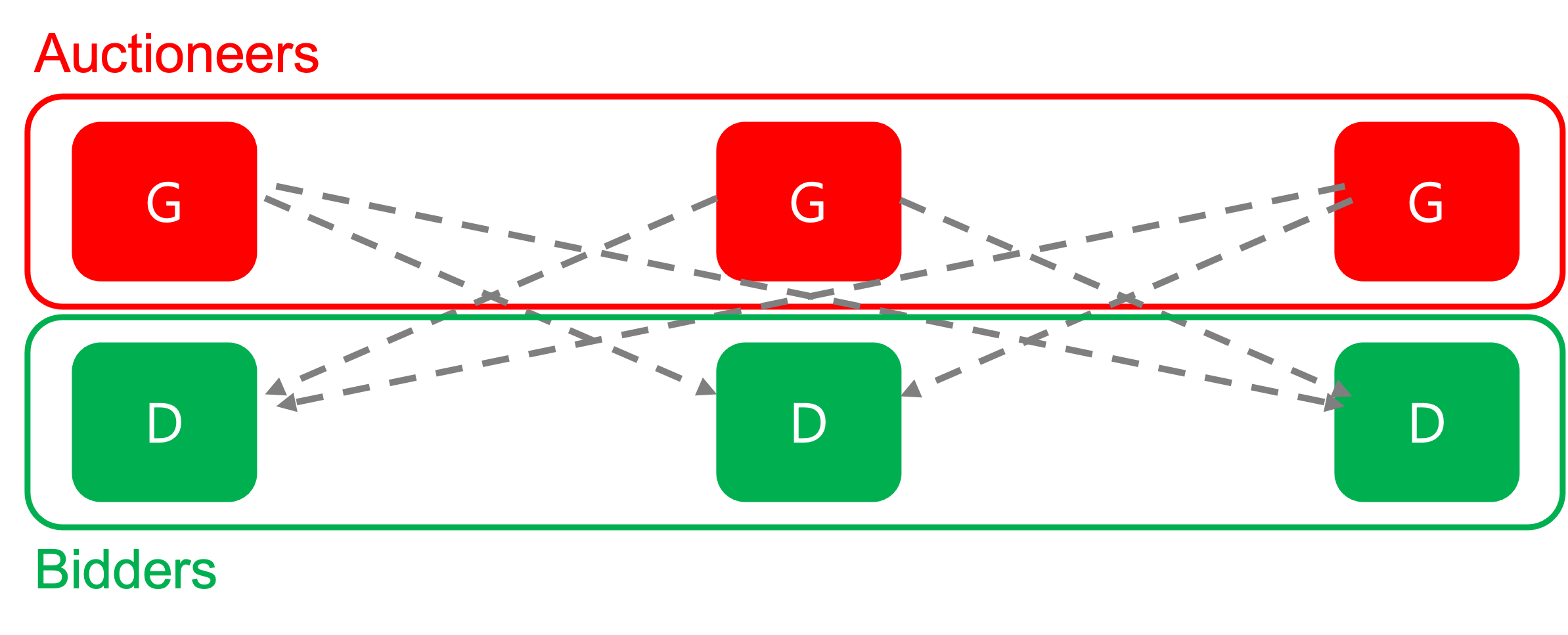}
    \caption{Example auction framework}
    \label{fig:Auction_framework}
\end{figure}

\begin{figure}[h]
    \centering
    \includegraphics[width=0.99\linewidth]{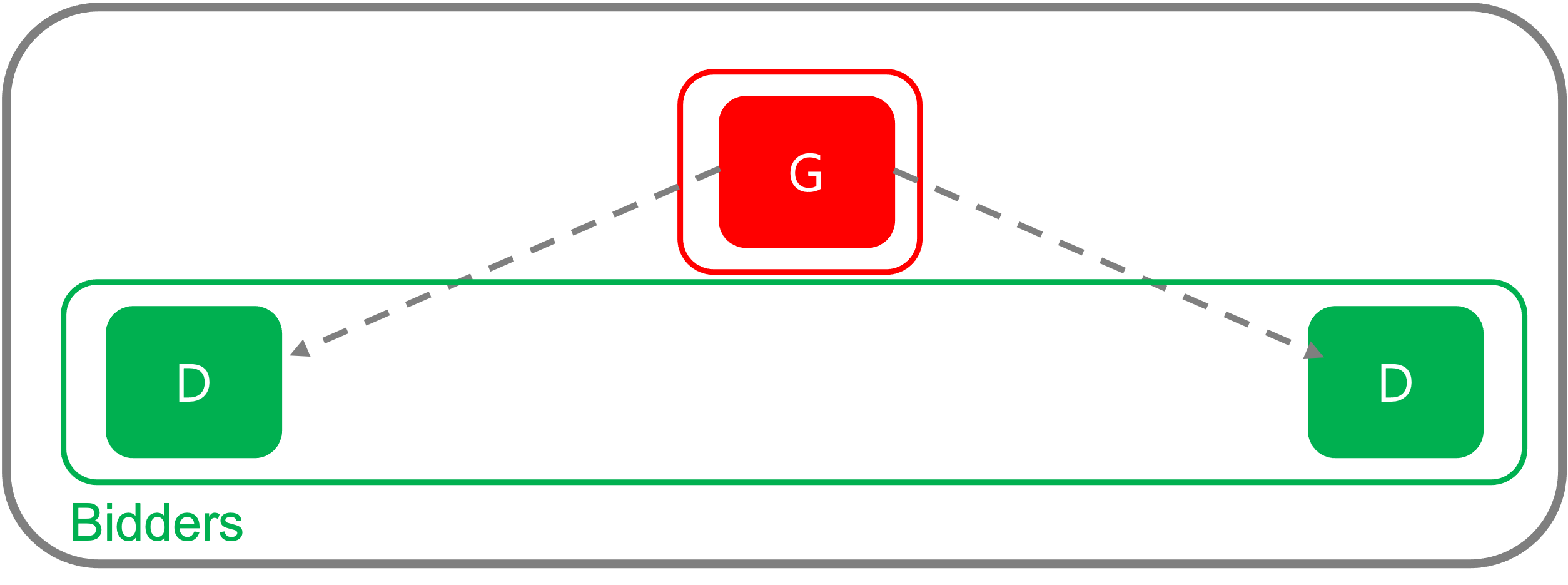}
    \caption{Example single auction phase}
    \label{fig:Auction_phase}
\end{figure}

Auctions are held for each generator (auctioneer), thus a total of $N$ auctions will be held. In each auction phase, all discriminators (participants), except their paired discriminator, produce a numerical value determining the authenticity of the generated images, and these values are employed as bidding values. The winner of each auction could be determined by using winner selection methods; nevertheless, the winner selection step is excluded as the main objective of this auction process is to attain accurate valuation for the generated images. An example of this procedure in the case of three pairs of GANs is illustrated in Figure.\ref{fig:Auction_framework} and Figure.\ref{fig:Auction_phase}.

The auction phase can be described as follows:
\begin{enumerate}
    \item The auctioneer $i$ presents a lot $L^i$, which is the group of $K$ items (samples) generated by the generator $G_i$
        \begin{align*} 
        L^i &= \{I_1^i,I_2^i,…,I_k^i\} \\ 
        I_k^i &\leftarrow G_i (z_k)
        \end{align*}
    
    \begin{figure}[h]
        \centering
        \includegraphics[width=0.8\linewidth]{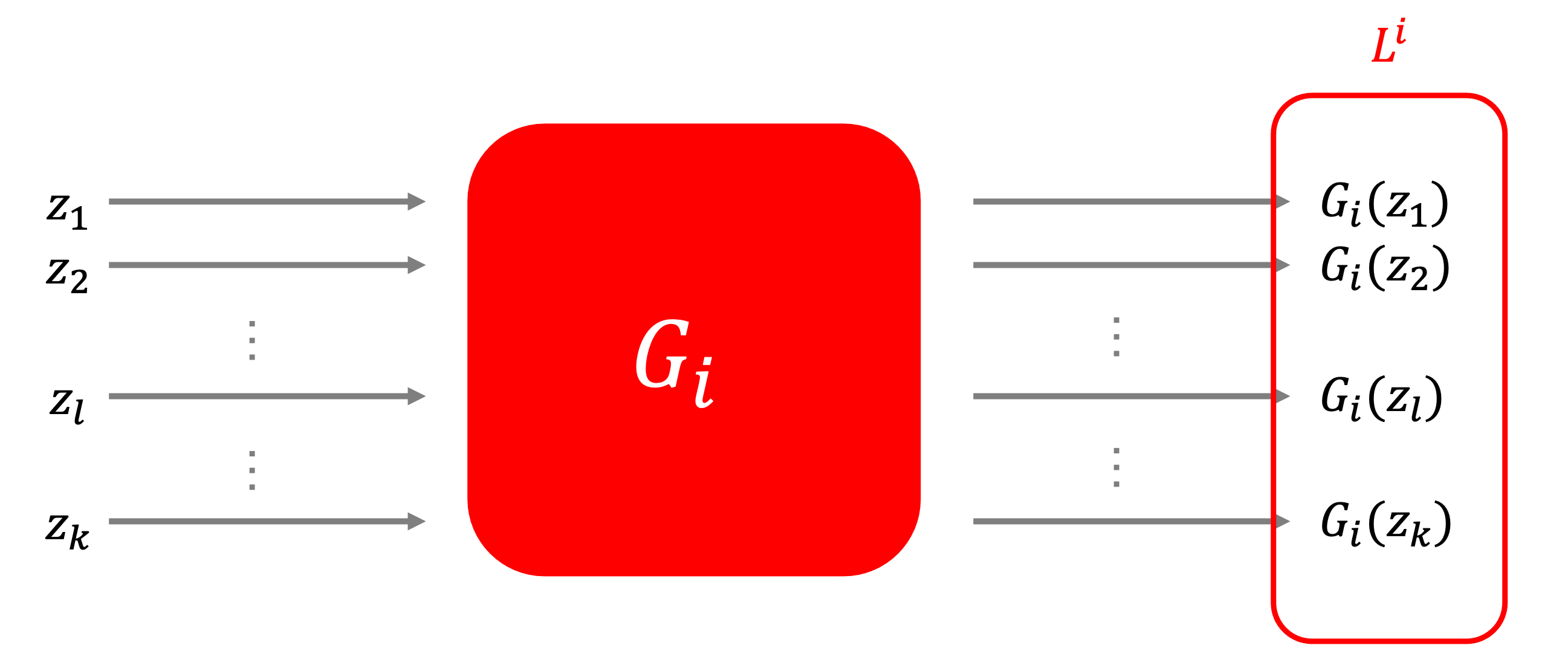}
        \caption{Auction process 1}
        \label{fig:A1}
    \end{figure}

    \item Each bidder $j$ ($j =1, …., N$, $j \neq i$) is asked to evaluate $K$ items from auctioneer $i$. The value of the item is the output of the discriminator $D_j$. The final bid $B^{ij}$ is the mean of values for all items. 
    \begin{align*} 
        B^{ij} &= \frac{1}{K} (B_1^{ij},B_2^{ij},…,B_k^{ij} )\\ 
        B_k^{ij} &\leftarrow D_j (I_k^i )
    \end{align*}
    \begin{figure}[h]
        \centering
        \includegraphics[width=0.99\linewidth]{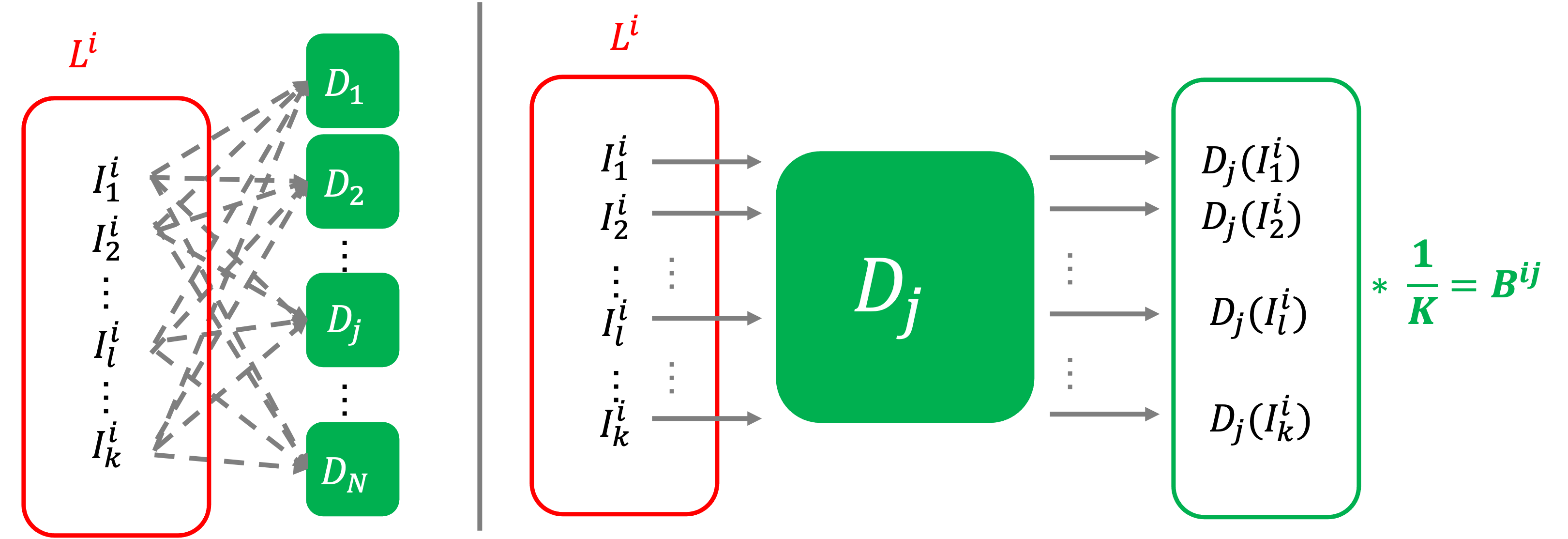}
        \caption{Auction process 2}
        \label{fig:A2}
    \end{figure}
\end{enumerate}

\subsection{Best GANs Selection}
Based on the auction result, the score of the GANs $i = (G_i, D_i)$ is determined by using the score function $S(i)$. This scoring involves calculating the mean difference of the bid prices received from all other participants for images generated by $G_i$, and the bid price submitted by $D_i$ for images presented by all other auctioneers. The score $S(i)$ is formulated as follows:
\begin{equation}
    S(i) = \frac{1}{N-1} \sum_{\substack{j=1 \\ j\neq i}}^{N} B^{ij}
    - \frac{1}{N-1} \sum_{\substack{j=1 \\ j\neq i}}^{N} B^{ji}
\label{eq:score}
\end{equation}
Here, $\sum B^{ij}$  is the sum of the values evaluated by all other discriminators $D_j (j\in {1,…., N}, j\neq i)$ for the images generated by $G_i$. This represents the sum of the probability values that discriminators of other GANs thought that the images generated by $G_i$ are from the real. Images generated by relatively poor GANs are more likely to be determined as fake images, that is, the value of $D_j(I_k^i)$ is more likely to be low, thus resulting in lower $B^{ij}$ values.  $\sum B^{ji}$  is the sum of the values that discriminator $D_i$ evaluated for the images generated by all other generators $G_j (j\in {1,…., N}, j\neq i)$. Similarly, this means the sum of probability values that $D_i$ thought as the real for the images generated by other generators. If the GANs $i$ has relatively poor performance, it is highly likely that images generated by other GANs $I_k^i \leftarrow G_i (z_k)$ will be judged as relatively good images, that is, $D_i(I_k^j)$ is more likely to be high. Therefore, the worse the GANs, the higher the value of $B^{ji}$ will be measured. Taken together, the score is designed as the above formula $S(i)$, and the better the performance, the higher the score.

\subsection{Auxiliary Loss}
The auxiliary loss is calculated using the determined best GANs $(G^*, D^*)$. 
The key idea is to match the loss output of a discriminator to the loss of the best discriminator $D^*$. 
Therefore, the auxiliary loss $L_{D^*}$ is defined as follows:
\begin{align}
    L_\text{aux}  = \text{MSE}( L_D(X,\bar{X}), L_{D^*}(X,\bar{X}) ),
\end{align}
where $L_D$ is the loss function of discriminator $D$.

Finally, optimize the objective augmented using the auxiliary loss:
\begin{equation}
    \mathcal{L}_D = L_D + \lambda  L_\text{aux},
\end{equation}
where $\lambda$ is a hyperparameter to be determined.

\section{Evaluations}
This section presents the analysis of the proposed algorithm applied to two classical GANs models: vanilla GANs \cite{goodfellow} and WGAN \cite{wgan}. The algorithm is demonstrated using a synthetic mixture of Gaussian distributions comprised of eight modes. A simple 2D-Gaussian dataset is used for the experiment because mode collapse is more evident in such a simple dataset. Specifically, eight GANs were trained using and without using the proposed algorithm in the same environment. For all experiments, ReLU networks containing two hidden layers each of width 256, are applied for all generators and discriminators. Additionally, 2-dimensional latent variables are used in all the experiments. 

\subsection{Qualitative Evaluations}
GANs trained using the proposed method cover various points without falling into mode collapse, while GANs simply trained in the standard way are often biased to one side or miss data points in some areas (see Figure \ref{fig:Bid_result_1}). In the classical training, some of the GANs showed biased results or holes in the data distribution, whereas all modes were covered where the proposed training method was applied. Although not as distinctly different as the results in GANs model, experiments on WGANs also show progressive results. WGANs trained using the proposed method covered various modes of data distribution in a balanced and sparse manner, while some samples were leaning on particular sides in WGANs where the proposed method was not applied (see Figure \ref{fig:Bid_result_2}).  

\subsection{Quantitative Evaluations}
Quantitative metrics are implemented to assess both the quality and the mode coverage of the generated samples. To evaluate quality, the mean likelihood estimation for modes is used. For the mode coverage assessment, the Wasserstein distance between the sample mode distribution and a uniform distribution is measured. In Figure \ref{fig:result_sparsity_gan}. and Figure \ref{fig:result_sparsity_wgan}., the mode coverage is compared quantitatively. For GANs models, it can be seen that the application of the proposed method is better at mode coverage. The higher the value, the higher the degree of spread of samples, which means there is less mode overfitting occurring. The differences between WGANs are not as clear as those of GANs, but there was slight progress in mode coverage. 

For the performance evaluations, the likelihood metric was used and there was an even slight improvement. All values in Table \ref{tab:data_info} and \ref{tab:data_info_w} are the mean values obtained from 10 experiments. The mean refers to the mean likelihood of eight GANs, and the min refers to the worst performance result among eight GANs. The performance of the worst has improved particularly, which is expected to be due to the auxiliary training that refers to the best GANs, resulting in preventing it from going to failure mode.

\begin{table}[h]
\centering
\caption{Likelihood Evaluations in GANs}
\bigskip
\label{tab:data_info}
\begin{tabular}{lcc}
\toprule
 & \textbf{classic} & \textbf{proposed}\\
\midrule
mean likelihood & $-3.530\pm0.556$ &  $-3.235\pm0.577$ \\
min likelihood &  $-6.414\pm1.528$ &  $-4.995\pm1.120$  \\
\bottomrule
\end{tabular}
\end{table}

\begin{table}[h]
\centering
\caption{Likelihood Evaluations in WGAN}
\bigskip
\label{tab:data_info_w}
%
\begin{tabular}{lcc}
\toprule
 & \textbf{classic} & \textbf{proposed} \\
\midrule
mean likelihood &  $-0.651\pm 0.831$ & $-0.478\pm0.571$ \\
min likelihood & $-4.756\pm 5.525$ & $-2.691\pm1.503$  \\
\bottomrule
\end{tabular}
\end{table}

\afterpage{
\begin{figure*}[t]
    \begin{minipage}{1\textwidth}
        \centering
        \includegraphics[width=1\linewidth]{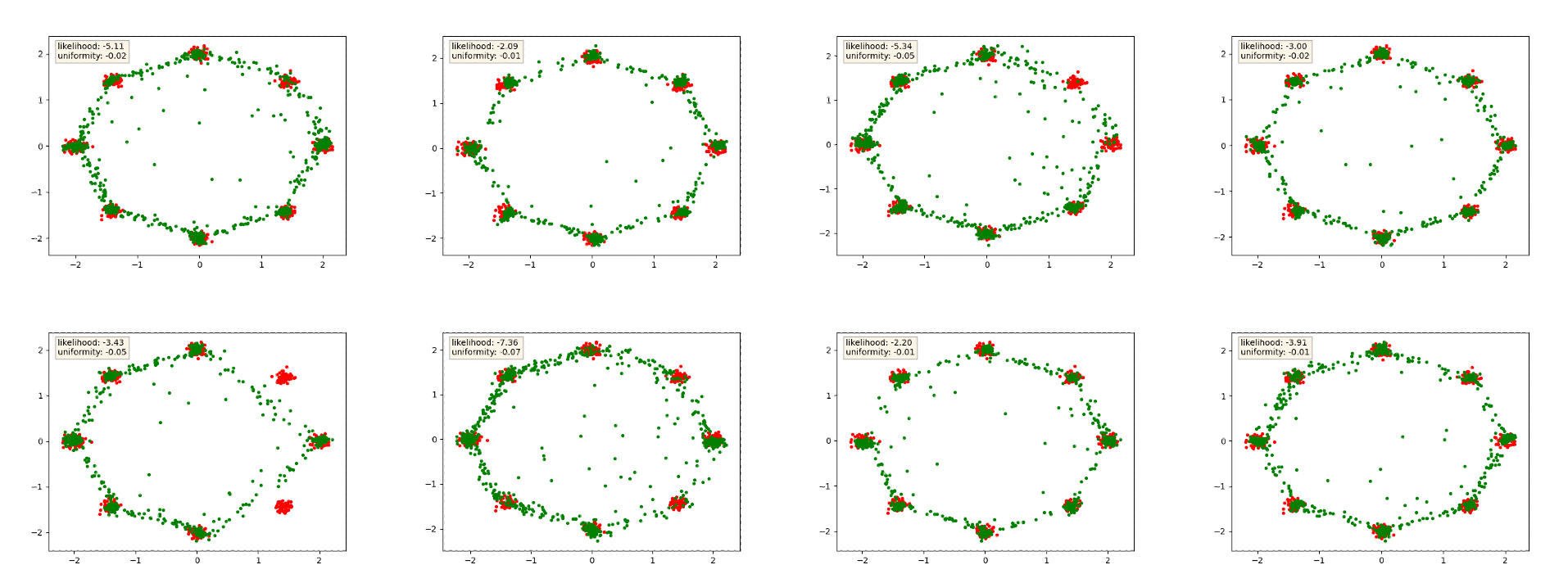}
        \label{fig:Bid_result_1_base}
    \end{minipage}%
    
    \hrule 

    \begin{minipage}{1\textwidth}
        \centering
        \includegraphics[width=1\linewidth]{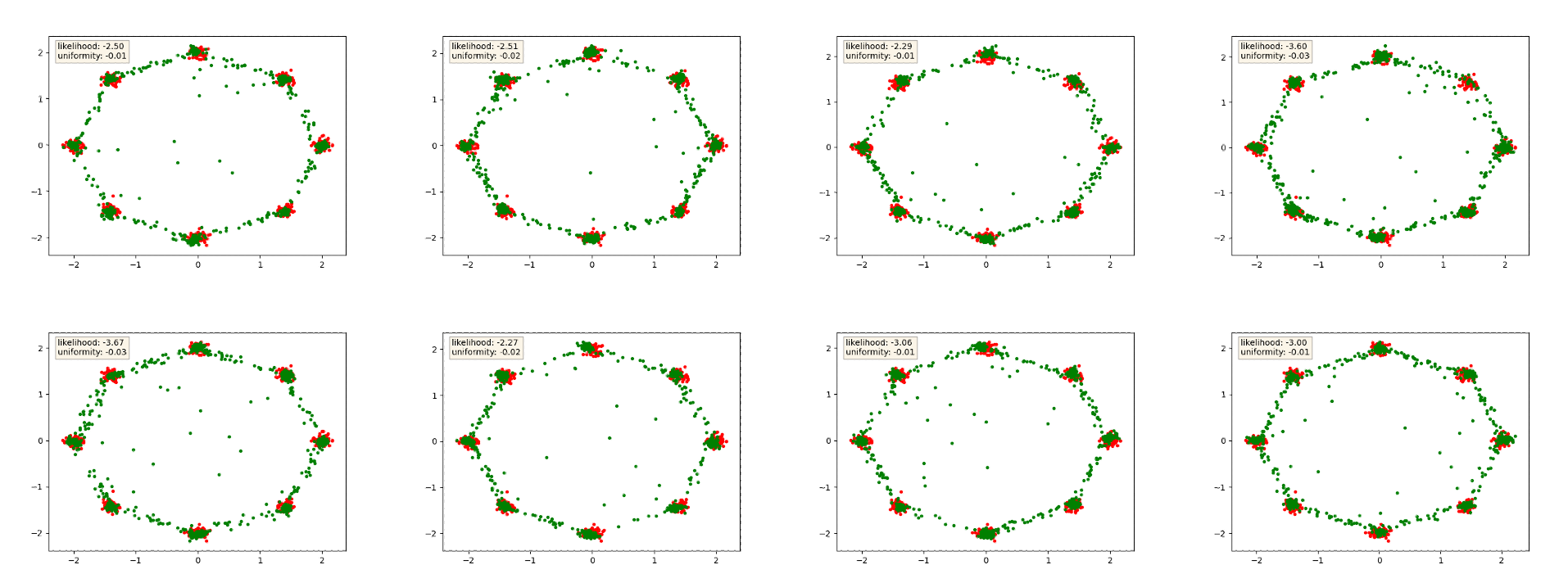}
        \label{fig:Bid_result_1_ours}
    \end{minipage}
    \caption{Qualitative evaluations using GANs. \\
        Top: Trained without proposed method; Bottom: Trained using the proposed method.}
    \label{fig:Bid_result_1}
\end{figure*}
\clearpage
}

\afterpage{
\begin{figure*}[t]
    \centering
    \begin{minipage}{1\textwidth}
        \centering
        \includegraphics[width=1\linewidth]{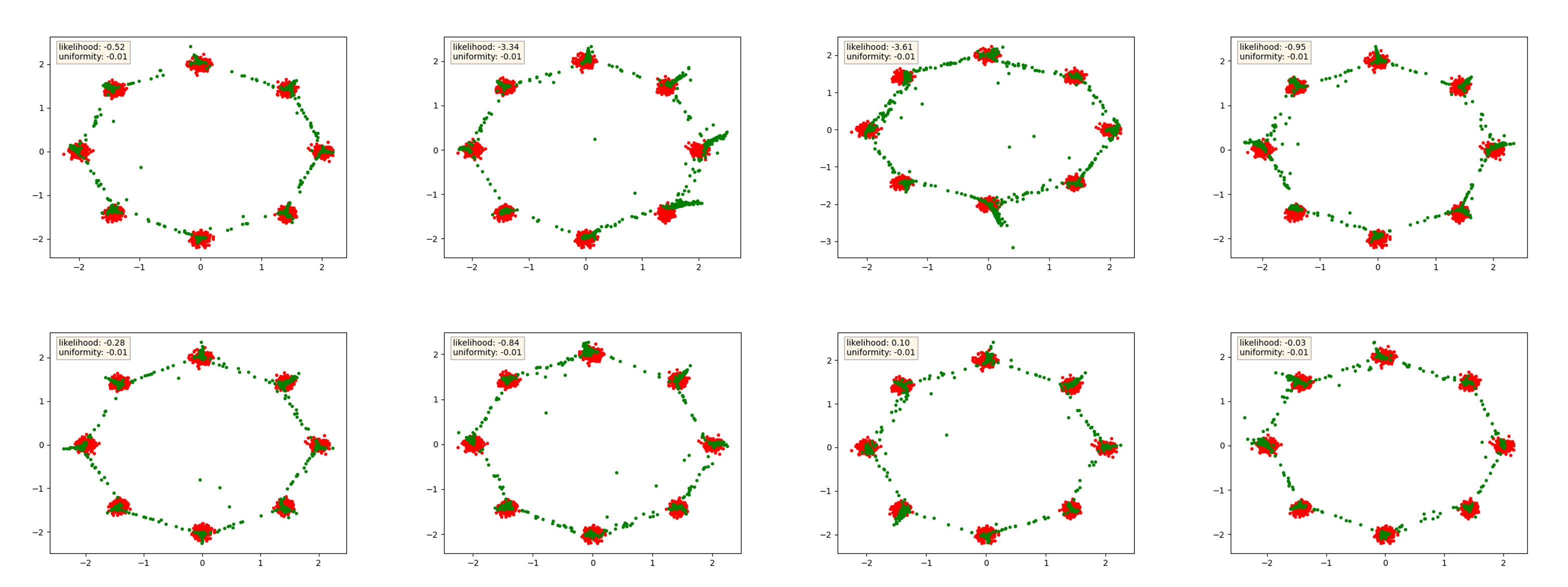}
        \label{fig:Bid_result_2_base}
    \end{minipage}%
    
    \hrule 

    \begin{minipage}{1\textwidth}
        \centering
        \includegraphics[width=1\linewidth]{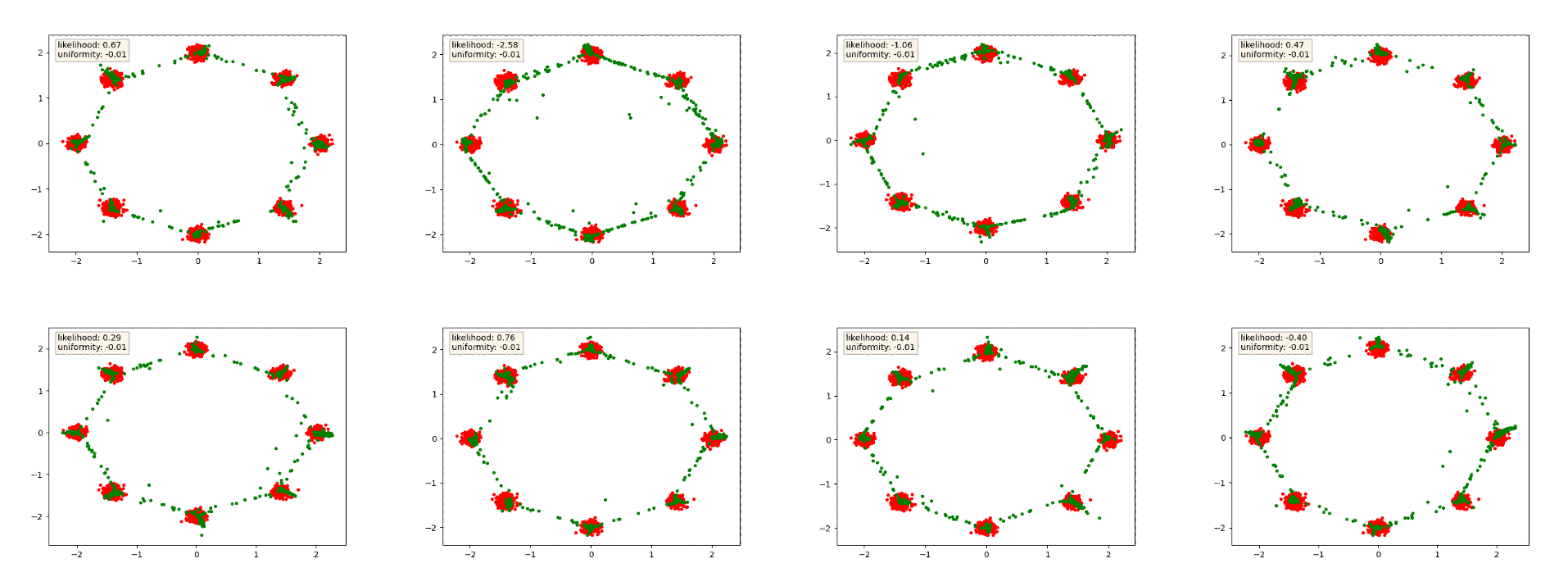}
        \label{fig:Bid_result_2_ours}
    \end{minipage}
    \caption{Qualitative evaluations using WGANs.\\ 
    Top: Trained without proposed method; Bottom: Trained using the proposed method.}
    \label{fig:Bid_result_2}
\end{figure*}
\clearpage
}

\afterpage{
\begin{figure*}[p]
    \centering
    \includegraphics[width=0.99\linewidth]{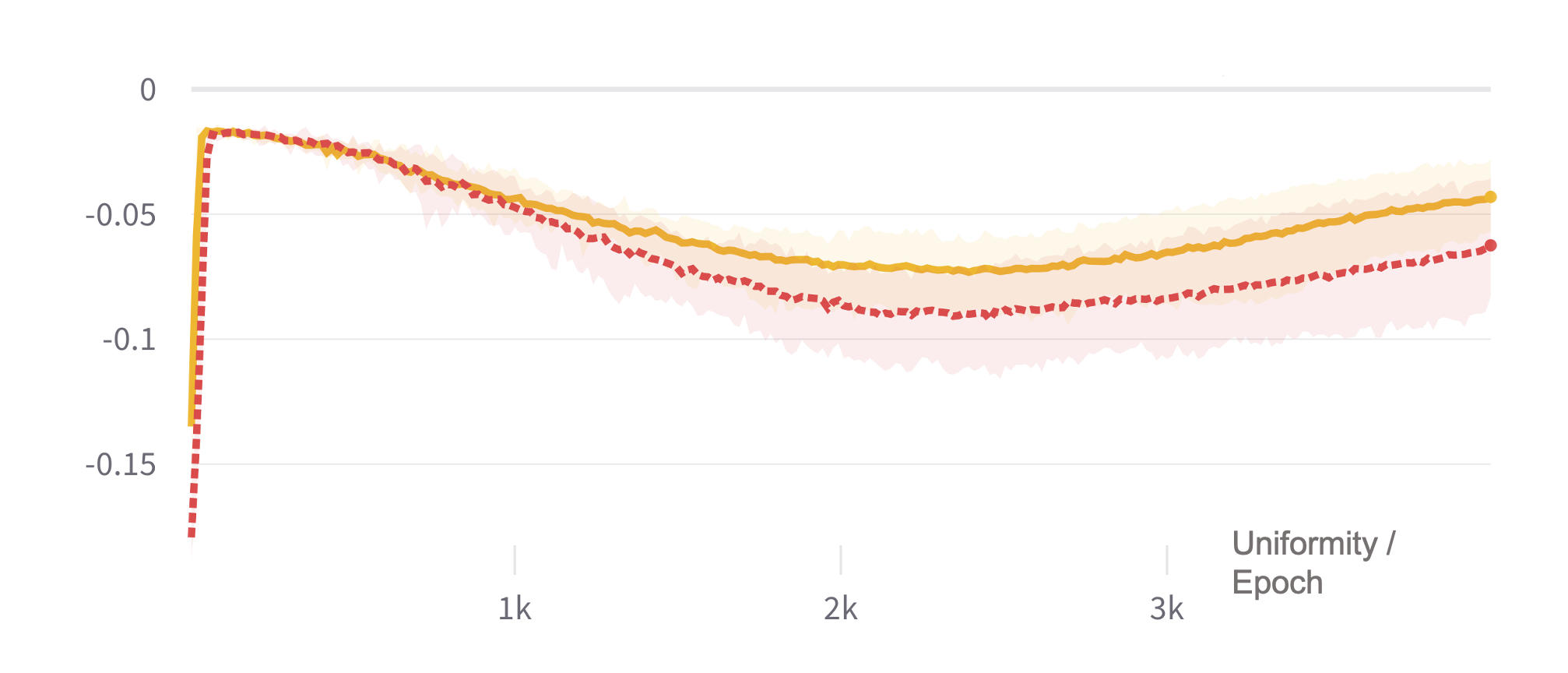}
    \caption[Quantitative evaluations using GANs]{Quantitative evaluations using GANs. yellow-line: trained without using the proposed method, red-dotted: trained using the proposed method}
    \label{fig:result_sparsity_gan}
\end{figure*}

\begin{figure*}[p]
    \centering
    \includegraphics[width=0.99\linewidth]{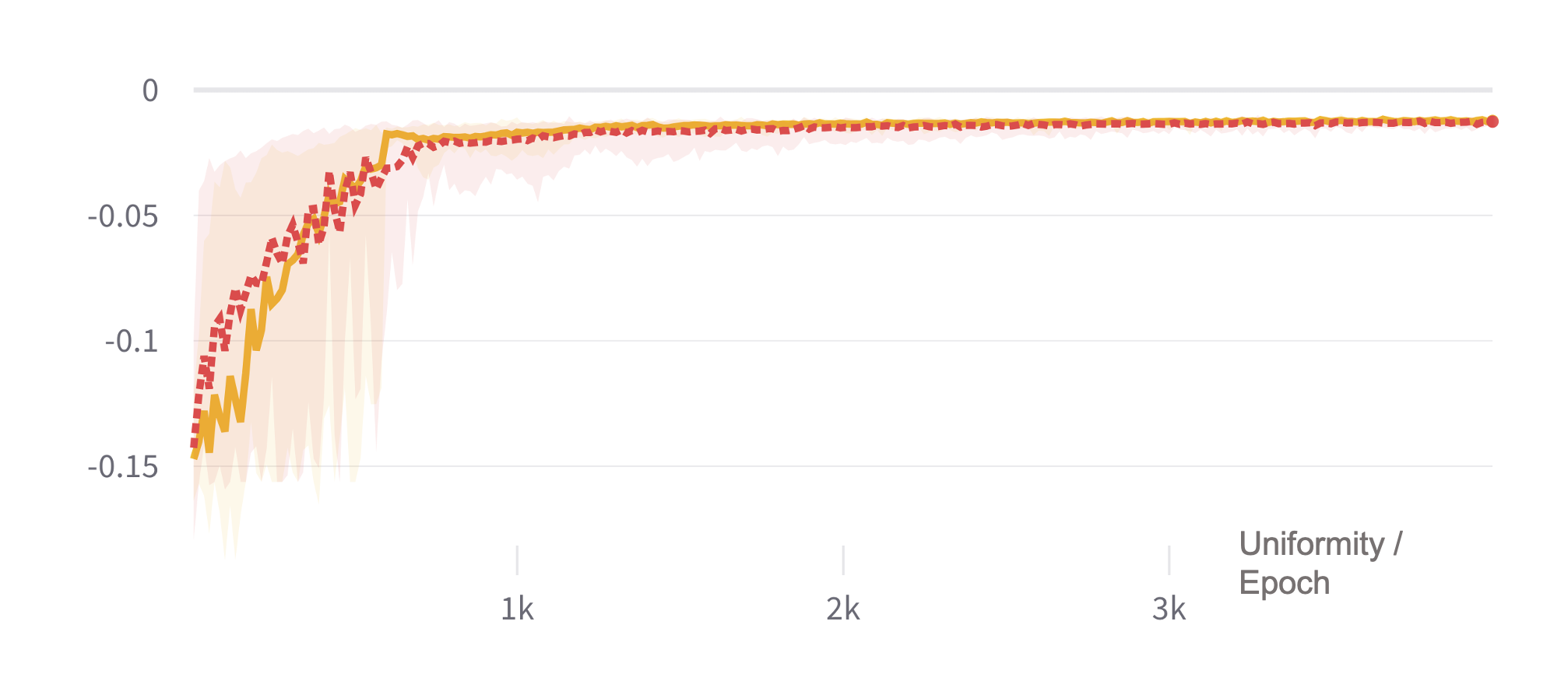}
    \caption[Quantitative evaluations using WGANs]{Quantitative evaluations using WGANs. yellow-line: trained without using the proposed method, red-dotted: trained using the proposed method}
    \label{fig:result_sparsity_wgan}
\end{figure*}
\clearpage
}

\section{Conclusion}
In conclusion, this work introduces a novel approach to tackle the mode collapse problem in GANs by extending the traditional two-player game into a multi-player game. This framework effectively trains multiple GANs, offering a solution to the mode collapse problem. One notable contribution of this work is the introduction of an auction-like evaluation approach, which assists in selecting and valuing the best-performing GANs. This approach guides the training process effectively and helps prevent failure modes. Additionally, an auxiliary training process is proposed, providing proper guidance for training multiple GANs.
While the proposed method has successfully mitigated the mode collapse and instability problem, it does require additional computational efforts. Future work will focus on improving performance and implementing the approach using other GANs models. Exploring better bid functions or alternative auction mechanisms may prove beneficial. Further investigation into updating using relative loss is also needed. All these limitations are left for future research. 

\bibliographystyle{IEEEtran}
\bibliography{generative_models,Bid,GANs}

\end{document}